# Taxi Dispatching Strategies with Compensations

Holger Billhardt[a], Alberto Fernández[a], Sascha Ossowski[a], Javier Palanca[b], and Javier Bajo[c]

[a] Centre for Intelligent Information Technologies (CETINIA), Universidad Rey Juan Carlos, Móstoles 28933, Madrid, Spain (e-mail: holger.billhardt@urjc.es, alberto.fernandez@urjc.es, sascha.ossowski@urjc.es).
[b] GTI-IA Research Group, Universitat Politècnica de Valencia, 40622 Valencia, Spain (e-mail: jpalanca@dsic.upv.es).
[c] Department of Artificial Intelligence, Universidad Politécnica de Madrid, Spain (e-mail: jbajo@fi.upm.es).

## Abstract

Urban mobility efficiency is of utmost importance in big cities. Taxi vehicles are key elements in daily traffic activity. The advance of ICT and geo-positioning systems has given rise to new opportunities for improving the efficiency of taxi fleets in terms of waiting times of passengers, cost and time for drivers, traffic density, $CO_2$ emissions, etc., by using more informed, intelligent dispatching. Still, the explicit spatial and temporal components, as well as the scale and, in particular, the dynamicity of the problem of pairing passengers and taxis in big towns, render traditional approaches for solving standard assignment problem useless for this purpose, and call for intelligent approximation strategies based on domain-specific heuristics. Furthermore, taxi drivers are often autonomous actors and may not agree to participate in assignments that, though globally efficient, may not be sufficently beneficial for them individually.

This paper presents a new heuristic algorithm for taxi assignment to customers that considers taxi *reassignments* if this may lead to globally better solutions. In addition, as such new assignments may reduce the expected revenues of individual drivers, we propose an economic compensation scheme to make individually rational drivers agree to proposed modifications in their assigned clients. We carried out a set of experiments, where several commonly used assignment strategies are compared to three different instantiations of our heuristic algorithm. The results indicate that our proposal has the potential to reduce customer waiting times in fleets of autonomous taxis, while being also beneficial from an economic point of view.

***Keywords*:** Coordination, dynamic fleet management, dynamic optimization, multi-agent systems, open systems, taxi assignment.

## 1. Introduction

Urban mobility is one of the main concerns that public managers face in big cities nowadays. Traffic congestions generate a high quantity of $CO_2$ emissions and cause extra time spent by travelers. One of the main actors involved in the daily traffic activity in urban areas are taxi fleets. They consist of several thousands of vehicles in big cities (e.g. about 15,000 taxis in Madrid, Spain). They are usually affiliated to mediator services, which coordinate service calls and taxi dispatching. Lately, new mobility systems that benefit from the advances in information and communication technologies have emerged, such as Uber[1], Lyft[2] or Liftago[3] among others.

Two of the main goals of a taxi fleet are (i) to reduce the response time (e.g., the time between a customer call and the moment a taxi arrives at the customer's location) and (ii) reduce costs of empty movements (e.g., movements taxis have to make in order to get to the location of customers). The provision of efficient methods for taxi assignment to customers is a challenge that can contribute to reducing distances of empty trips with the resulting decrease of traffic flow, pollution, time and so on. Typically, taxi fleet coordination companies apply the *first-come first-serve* strategy to assign taxis to customers. Once the taxi accepts the passenger, the dispatching is irreversible. This method is known to be inefficient (Egbelu & Tanchoco, 1984).

---
[1] http://www.uber.com
[2] http://www.lyft.com
[3] http://www.liftago.com

The aforementioned case falls into a specific class of assignment problems which is characterized by a dynamic demand in time and space. To efficiently solve such problems, dynamic algorithms are required instead of classical assignment optimization methods. For this purpose, techniques from the field of intelligent systems are promising, because they allow for developing heuristics-based algorithms that intelligently prune the search space, so as to reduce the computational complexity and to support a sufficient degree of scalability. Furthermore, taxi drivers are usually autonomous actors, i.e. they can freely choose whether to accept or to reject a recommendation proposed by the mediator service, which puts additional constraints on the set of feasible solutions to the assignment problem. As Ossowski and Omicini (2002) argue, dynamic coordination problems with self-interested actors can be effectively modelled as multiagent systems. Agreement Technologies (Ossowski et al. 2013) refer to a sandbox of methods within the field of multiagent systems enabling knowledge-based software agents to interact with each other so as to forge agreements on behalf of their users. Multiagent interaction protocols based on the algorithm first proposed by Bertsekas (1984), for instance, coordinate software agents by iteratively simulating auctions among them, and have been successfully applied to a dynamic assignment problem in the domain of emergency management (Billhardt et al., 2018).

In this article, we deal with the problem of dynamic taxi assignment to customers with the goal of minimizing the global waiting time of passengers. Our heuristic assignment algorithm considers taxi *reassignments* if this may lead to globally better solutions. That is, taxis that have been dispatched to pick up a customer but are still on their way may be reassigned to another customer. For this purpose, we adapt the method put forward by Billhardt et al. (2014) for ambulance management to the taxi assignment problem. In addition, we go beyond that approach by taking into account the taxi drivers' autonomy: in case of an assignment change that improves the efficiency of the taxi fleet at global level but may be disadvantageous for some individual taxi driver (e.g., she may be assigned a customer located further away compared to the initially assigned one), that driver will receive a compensation to make the assignment agreeable to her as well. To the best of our knowledge, there are no other approaches that consider reassignment until pick-up time and taxi autonomy to propose a new scheduling when new customers show up or taxis become available.

The main contribution of this paper is twofold. Firstly, we introduce an algorithm for taxi dispatching, which exploits dynamic taxi reassignment and is coupled with an economic compensation schema which assures that (rational) taxi drivers will freely accept reassignments of "worse" customers if this can improve the overall performance of the system. Secondly, we perform an evaluation of three instances of the proposed algorithm, minimizing or maximizing different parameters (distance, revenues, and a combination of both).

The rest of the paper is organized as follows. Section 2 analyzes existing works related to taxi assignment. In section 3, we describe the problem we are dealing with and some common dispatching strategies. In section 4 the proposed taxi reassignment algorithm and compensation schema is described. Section 5 details the experiments carried out to evaluate our approach. Finally, we conclude the paper with section 6.

## 2. Related Work

The development of ICT, especially GPS and wireless connectivity, has driven the proposal of many taxi assignment systems during the last decade.

Many works are centered analyzing new assignment strategies in order to reduce the waiting times of customers. The classical approach is the *first-come/first-served (FCFS)* strategy, where each new customer is assigned to the nearest available taxi. Lee, Wang, Cheu, and Teo (2004) present a system that takes advantage of real-time information (on taxis and traffic conditions) to assign taxis with the shortest time path to customers, instead on the closest taxis. Maciejewski, Salanova, Bischoff, and Estrada (2016) compare the classical FCFS strategy with a *demand-supply balancing strategy* that assigns taxis to the closest customers in high demand scenarios (instead of customers to taxis) in microscopic simulations of taxi services in Berlin and Barcelona. In (Maciejewski, Bischoff, & Nagel, 2016), they present a real-time dispatching strategy based on solving the taxi assignment problem. In this approach, the assignment problem is considered from a more global perspective. They propose to calculate the optimal assignment among idle taxis and pending requests at certain intervals or whenever new events (new customer/available taxi) take place. Zhu and Prabhakar (2017) analyze how suboptimal individual decisions lead to global inefficiencies and propose an assignment model based on network flow.

While most existing approaches try to minimize the average waiting time of customers, other works have a different focus (Dai, Huang, Wambura, & Sun, 2017; Gao, Xiao, & Zhao, 2016; Meghjani, & Marczuk, 2016; and Ngo, Seow, & Wong, 2004). BAMOTR (Dai et al., 2017) provides a mechanism for fair assignment of drivers, where fair assignment is intended to minimize the differences in income among the taxi drivers. For that, they minimize a combination of taxi income and extra waiting time. Gao et al. (2016)

propose an optimal multi-taxi dispatching method with a utility function that combines the total net profits of taxis and waiting time of passengers. They also consider different classes of taxis. Meghjani and Marczuk (2016) propose a hybrid path search for fast, efficient and reliable assignment to minimize the total travel cost with a limited knowledge of the network. In (Ngo et al., 2004), a fuzzy approach is proposed for defining the cost function to be minimized, which encompass a fuzzy aggregation of multiple vague criteria defined by human experts.

There are other works that focus on taxi demand prediction with the goal of helping taxis to quickly find closer passengers or of balancing supply and demand of taxis in an area of interest. Grajciar (2015) proposes a method for recommending areas where idle taxis are more likely to find a new customer. The price of each journey is not fixed and is proposed to the customer by a broker. Then, taxis bid for the customer. Moreira-Matias et al. propose methods for predicting taxi-passenger demand (Moreira-Matias, Gama, Ferreira, Mendes-Moreira, & Damas, 2013) and profitability (Moreira-Matias, Mendes-Moreira, Ferreira, Gama, & Damas, 2014) at taxi stands. Zhang et al. (2015) model the taxi driver's service strategies from three perspectives: passenger-searching, passenger-delivery, and service-area preference. Miao et al. (2016) and Miao et al. (2017) treat the problem of dispatching vacant taxis towards current and future demands while minimizing total idle mileage. Their approach is based on forecasting the uncertain spatial-temporal taxi demands in a region.

There are also an increasing number of works on taxi ridesharing (e.g., Ma, Zheng, & Wolfson, 2013; d'Orey, Fernandes, & Ferreira, 2010; Li, Horng, Chen, & Cheng, 2016; Tian, Huang, Liu, Bastani, & Jin, 2013; and Mareček, Shorten, & Yu, 2016), although in this paper we do not focus on that problem.

In our work we concentrate on the problem of dispatching (assigning) taxis to (current) customers. In contrast to other works in this field, the main characteristic of our approach is that we treat the problem from a global and dynamic perspective. In particular, we try to find assignments from taxis to pending customers that globally minimize the expected waiting times of customers. Furthermore, we consider the possibility of modifying an existing assignment when a taxi has been dispatched but has not yet picked up the corresponding customer. We already successfully followed a similar approach in our previous work on taxi assignments (Billhardt et al., 2017), and in (Billhardt et al., 2014) to assign ambulances to patients in emergency medical services. One of the few other works in this line is (Glaschenko, Ivaschenko, Rzevski, & Skobelev, 2009), which presented an adaptive scheduling in which reassignment is possible during a time interval until pick-up order is sent to the taxi and customer. During this process, vehicle agents negotiate with each other. In our case, we do not restrict reassignment to a specific interval. Furthermore, since modifications of existing assignments may imply changes in the expected incomes of a taxis, we propose a method that economically compensates taxis such that modifications in their current assignments will not result in a loss of income.

## 3. Problem Definition and standard Dispatching strategies

In this section we describe in more detail the problem we are tackling in this article. Table 1 contains a list of symbols used in the rest of the paper.

Table 1. List of symbols used in equations.

| Symbol | Meaning |
|---|---|
| $T$ | Set of taxis and $i \in T$ denotes a taxi |
| $T^D$ | Set of assigned (i.e. dispatched) taxis |
| $T^O$ | Set of occupied taxis |
| $T^A$ | Set of available taxis |
| $C$ | Set of customers and $k \in C$ denotes a customer |
| $C^A$ | Set of customers assigned to a taxi |
| $C^U$ | Set of customers waiting to be assigned to a taxi |
| $C^S$ | Set of customers in service (in a taxi) |
| $D_k$ | Destination of customer $k$ |
| $d_{ik}$ | Distance from taxi $i$ to the location of customer $k$ |
| $t_{ik}$ | Time it takes taxi $i$ to reach customer $k$ |
| $d_{kd}$ | Distance from pick up location of customer $k$ to his destination $d$ |
| $d_{ikd}$ | Total distance to serve customer $k$ with taxi $i$ ($d_{ik} + d_{kd}$) |

We consider taxi systems in which there exist some mediator service in charge of coordinating the assignment of customers to taxis. Customers contact the mediator via phone calls or any other telematic means available nowadays. The mediator dispatches taxis to serve customers. We do not explicitly deal with group of customers. If several people travel together they are considered as one customer (e.g. the one that made the contact).

Customer $k$, requests a taxi at time t. The mediator assigns taxi $i$ to serve customer $k$. Then, taxi $i$ moves to location of $k$, with driving distance $d_{ik}$, which takes $t_{ik}$ time to reach $k$. After picking up $k$ the taxi drives to the customer's destination $D_k$, located at distance $d_{kd}$. We denote by $d_{ikd} = d_{ik} + d_{kd}$ the total distance for serving customer $k$ with taxi $i$.

Let $T$ be the set of all taxis. Each taxi $i \in T$ can be either *available* (neither assigned nor occupied), assigned (*dispatched*, on the way to pick up a customer) or *occupied* (carrying customers), denoted by $T^A$, $T^D$ and $T^O$, respectively, and such that:

$$T = T^A \cup T^D \cup T^O \text{ and } T^A \cap T^D \cap T^O \equiv \varnothing$$

Let $C$ be the set of customers in the system at a given time. Each customer $k \in C$ can be either in *service* (inside a taxi), *assigned* to a taxi (a taxi has been dispatched to pick up the customer) or *unassigned* (waiting to be assigned to a taxi), denoted by $C^S$, $C^A$ and $C^U$, respectively, and such that:

$$C = C^S \cup C^A \cup C^U \text{ and } C^S \cup C^A \cup C^U \equiv \varnothing$$

The taxi assignment problem consists in dispatching customers, that is, assigning customers in $C^U$ to available taxis in $T^A$. The general goal of a dispatching mechanism is to optimize the existing resources and to reduce the waiting time of customers, that is, the time or distance it takes taxis to reach their customers (independently of their destination). In this paper, we assume time is proportional to distance, so we concentrate on reducing distances.

**Definition** (*Taxi Assignment*). Given a set of taxis $T$ and a set of customers $C$, a taxi assignment $A$ is a set of pairs $<i,k>$ where $i \in T$ and $k \in C$, and typically such that all customers are assigned to a taxi (if the number of customers is lower than the number of taxis) or all taxis are assigned to a customer (in the other case). In addition, a customer cannot be assigned to more than one taxi and a taxi cannot be assigned to more than one customer.

The most common dispatching strategy is *first-come/first-served (FCFS)*, in which any customer in $C^U$ who is waiting the longest is assigned to the nearest available taxi from $T^A$. The assignment process is repeated whenever a new customer requests a taxi or whenever a taxi becomes available after a previous trip.

The FCFS strategy always dispatches to each unassigned customer the nearest taxi. However, in very high demand scenarios, e.g., when the number of taxis is lower than the number of customers ($|T^A| < |C^U|$), this strategy turns out to perform quite badly (Maciejewski et al., 2016). An alternative that solves this problem is the *nearest-taxi/nearest-request (NTNR)* strategy. In NTNR customers are assigned in the same way as in FCFS if the number of unassigned customers ($|C^U|$) is lower or equal to the number of available taxis ($|T^A|$). However, if $|T^A|<|C^U|$, the assignment is processed on the side of the taxis, assigning each available taxi to the closest customer in $C^U$. The rationale behind this approach is that in high demand scenarios, the overall waiting times of customers can be reduced if the closest customers are served first.

Dispatching strategies like FCFS and NTNR, do not optimize the assignments globally and do not take into account possible improvements of a global assignment at a given point in time that might exist because of the dynamic nature of taxi services. These methods do not allow for a reassignment of already dispatched taxis which could in certain occasions achieve better results. This can be seen in the example shown in Figure 1. Here, a scenario with two taxis is presented ($t_1$ and $t_2$). The first customer in appearing is $c_1$, and $t_1$ is dispatched to pick it up since it is located closer (1.8 km) than $t_2$ (2 km). A few minutes later a new customer $c_2$ appears. At this moment FCFS and NTNR would assign $t_2$ to $c_2$. The overall travel distance towards the customers (at this moment) would be 4km. However, there is a better global assignment $A'=\{<t_1,c_2>,<t_2,c_1>\}$ with a total of 3.5km.

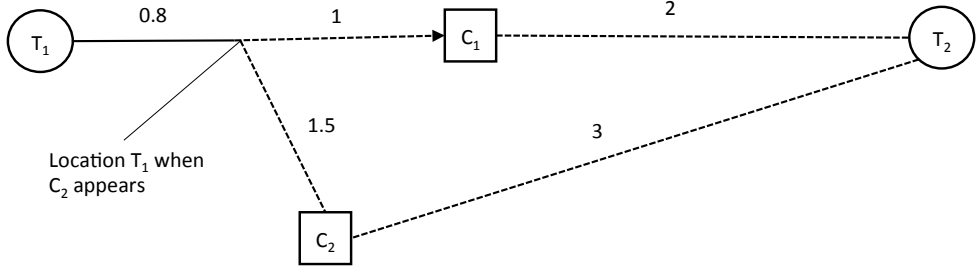

**Figure 1.** Example of scenario where a new customer appears. Numbers indicate distance in km. The solid line indicates the distance driven by $T_1$ when $C_2$ appears.

A way to reduce overall travel distances of taxis at a given point of time, consists in allowing reassignment of customers to taxis and finding the globally best assignment of all available or already dispatched taxis ($T^A \cup T^D$) to all unassigned or assigned customers ($C^A \cup C^U$). This can be done by solving the *assignment problem* (Bertsekas, 1988). In our previous work (Billhardt et al., 2017), we applied this idea using Bertsekas' auction algorithm to find optimal assignments. The optimization process is applied whenever the situation changes and a new optimal solution may exist, that is, whenever a new customer appears or a previously occupied taxi becomes available again (or starts working).

In this way it is assured that the assignment of taxis to customers is optimal at each moment (with respect to the global distances of taxis to customer' locations). We call this dispatching strategy *Full auction*[4] *(FA)* (for details the interested reader is referred to (Billhardt et al., 2017)).

## 4. Taxi Reassignments with Compensation

The reassignment strategy FA will be appropriate in scenarios of taxi fleets operated by taxi companies and where the taxi drivers have a fixed salary that does not depend on the trips they are doing. However, the strategy will not work for fleets of autonomous taxi drivers for which their revenues depend on the customers they serve during the day. This is basically due to the fact that autonomous taxi drivers will not accept a reassignment of customers if this would imply a reduction in their income.

The idea of the proposal we present in this section is to define a dispatching strategy that involves the reassignment of taxis, but would always be accepted by autonomous taxi drivers (who rationally decide based on maximizing their profit).

### 4.1. Taxi revenues and the effect of a taxi reassignment

To analyze the revenue of a taxi for a trip we assume the following payment and cost scheme. Customers pay a fixed cost *fcost* per trip plus a *fare* per kilometer for the distance from their pickup location to their destination. Furthermore, a taxi has a *cost* per kilometer for car usage (including petrol and other expenses, e.g. maintenance, taxes, etc.). With this structure, the monetary revenue of a taxi $i$ for serving customer $k$ is:

$$Revenue(i,k) = fcost + fare \cdot d_{kd} - cost \cdot d_{ikd},$$

where $d_{ikd}$ and $d_{kd}$ are the total distance and the distance from $k$ to its destination $D_k$, respectively.

When a taxi $i$, previously assigned to a customer $k$ in assignment $A^o$ is reassigned to customer $j$ in a new assignment $A^{new}$, as presented in Figure 2, its revenue would change by:

$$\Delta Revenue(i,k,j) = Revenue(i,j) - Revenue(i,k) = fare \cdot (d_{jd} - d_{kd}) + cost \cdot (d_{ikd} - d_{ijd})$$

The economic effect of the new assignment on i (reassigned from $<i,k> \in Ao$ to $<i,j> \in Anew$) can be zero, positive or negative.

---

[4] The algorithm is based on an auction process

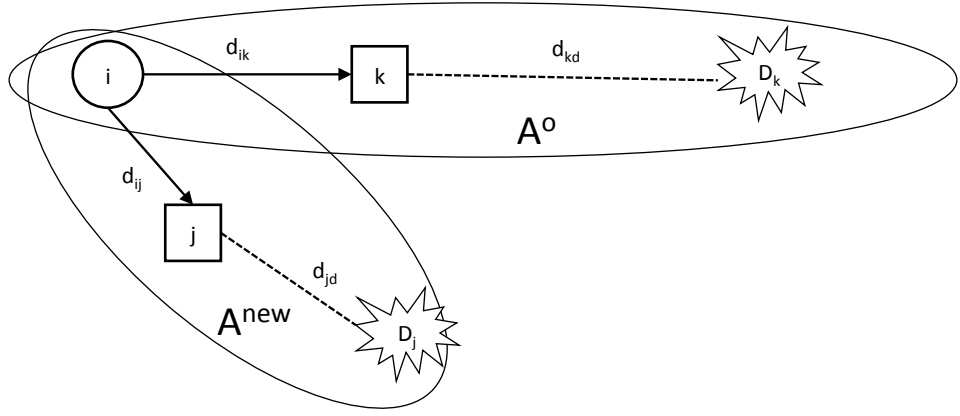

**Figure 2.** Example of reassignment of taxi $i$ from customer $k$ in $A^o$ to $j$ in $A^{new}$.

## 4.2. Reassignment compensations

We assume taxi drivers to be economically rational, that is, they want to maximize their income and minimize the time spent on their trips. In particular, we make the following assumptions:
   a) A taxi driver that is available will always accept a new customer.
   b) A taxi driver would always prefer earning the same amount of money in less time.
   c) A taxi driver would always accept any extra movement (with a distance $d$) if this implies an extra income of $d \cdot (fare - cost)$. In fact, this is actually the current rate for which a taxi driver is working.

Based on these assumptions we define a compensation $c$ that is applied if a taxi $i$ accepts a reassignment from customer $k$ to $j$. Taking into account this compensation, the effective revenue of taxi $i$ when accepting customer $j$ would be $Revenue'(i,j) = Revenue(i,j) + c$. We consider the following two situations:
   a) If $d_{ikd} < d_{ijd}$:
        $c = Revenue(i,k) - Revenue(i,j) + (d_{ijd} - d_{ikd}) \cdot (fare - cost)$ and thus
        $Revenue'(i,j) = Revenue(i,k) + (d_{ijd} - d_{ikd}) \cdot (fare - cost)$
      Here, the taxi would be compensated for the extra distance with the standard fare.
   b) If $d_{ikd} \geq d_{ijd}$:
        $c = Revenue(i,k) - Revenue(i,j)$ and thus
        $Revenue'(i,j) = Revenue(i,k)$
      In this case, the taxi would earn the same as before, but for a service with the same or a shorter distance than the previous one.

It is clear that, with the assumptions on economic rationality of taxi drivers, a taxi would accept any reassignment with the defined compensations.

It should be noted that compensations may be positive or negative, i.e., a taxi may receive a payment from the mediator in addition to the fare that it collects from the client, but it may also have to pay to the mediator part of the fare that it charges to client. For instance, if the customers' destinations are unknown, the distances $d_{-d}$ may need to be estimated by some constant which is the same for all customers. Then, in the above case (b), the compensation would be:
   $c = Revenue(i,k) - Revenue(i,j) = cost \cdot (d_{ij} - d_{ik})$

Since in this case, $d_{ik} > d_{ij}$, the taxi would have to pay to the mediator the cost of the difference in distances towards the new customer wrt. the previous customer.

Taxi compensations can be managed by the mediator entity, which is in charge of proposing new assignments as well as collecting and paying compensations to affected taxis. In this sense, a change from one global assignment $A^o$ to a new assignment $A^{new}$ implies revenues/cost for the mediator.

**Definition** (*Mediator Revenues*): Given a current assignment $A^o$ and a new assignment $A^{new}$, the economic effect on the mediator when applying $A^{new}$ with compensations, denoted as $R(A^o, A^{new})$, is defined as the sum of payments received from taxis minus compensations given.

## 4.3. Proposed algorithm

Algorithm 1 shows our proposal for a dispatching strategy based on reassignments with compensations. Similar to the FA strategy, the algorithm is executed whenever at least a new customer appears or a taxi becomes available.

The algorithm is in several ways influenced by the fact that we want to assure the mediator revenues to be positive over time, that is, we want a mediator that has no extra cost or even obtains some profit.

The first step (line 3) is to assign possible unassigned customers ($C^U$) to available taxis ($T^A$) using the NTNR strategy. The result ($A^o$) is a valid assignment that could be applied. However, we try to find a better assignment $A'$ (line 4) considering all pending (if any) and assigned customers ($C^A \cup C^U$) and only assigned taxis ($T^D$). This implies, that when finding assignment $A'$, we only consider the reassignment of customers among taxis that already have a customer assigned; in other words, no taxi would lose a customer. This step assures that the necessary compensation the mediator would have to pay to taxis are rather small. If we would allow de-assignments of customers (e.g., leaving previously assigned taxis unassigned in $A'$), the compensation costs for the mediator would be generally to high leading to the case that the mediator would imply extra cost.

**Algorithm 1.** Taxi Assignment

```
1:   Input: current assignment A^current, accumulated mediatorRevenue
2:   Output: assignment A^new
3:   A^o = A^current ∪ NTNR(T^A, C^U)
4:   A' = Calculate optimal assignment from (C^A∪C^U) to T^D
5:   for all <i,j> ∈ A' \ A^o do
6:       if d_ikd ≥ d_ijd | <i,k> ∈ A^o then
7:           c = Revenue(i,k) – Revenue(i,j)
8:       else
9:           c = Revenue(i,k) – Revenue(i,j) + (d_ikd – d_ijd)·(fare – cost)
10:      end if
11:      mediatorRevenue – = c
12:  end for
13:  if mediatorRevenue ≥ 0 then
14:      A^new = A'
15:  else
16:      A^new = A^o
17:  end if
18:  return A^new
```

Each modification of taxi $i$ is analyzed in lines 5-12. We calculate the required compensation $c$ for each taxi and add $-c$ to the accumulated mediator revenue. The new assignment is only accepted (proposed to the taxis) if the mediator keeps a positive accumulated revenue (lines 13-18).

We have implemented three different optimization functions (line 4):

- Minimizing distances to customers (*MinDist*): The optimization consists in finding the assignment that maximizes the reduction in the sum of distances of all dispatched taxis towards customers, with respect to the current assignment.
- Maximizing mediator revenue (*MaxRev*): The optimization consists in finding the assignment that maximizes the outcome of the mediator. The current assignment would have an outcome of 0 and, thus, acts as a lower limit.
- Minimizing distances /Maximizing mediator revenue (*MinDist/MaxRev*): a linear combination of the previous methods. Let $A$ be the current assignment and let $D(A)$ be the sum of the distances of taxis towards the customers in $A$. This assignment approach computes an assignment $A'$ that minimizes: $D(A') - \gamma \cdot R(A,A')$, where $\gamma$ is a ratio of scaling monetary income into distance values (meters) (e.g. the net benefit a taxi receives per meter when transporting a passenger). Therefore, this method optimizes two characteristics: it minimizes the distance to the assigned customers and it maximizes the revenue of the mediator.

## 5. Evaluation

In order to evaluate the effectiveness of our proposal we tested it in different experiments simulating the operation of a taxi fleet in an area of about 9×9 km, an area that roughly corresponds to the city center of Madrid, Spain. In the simulations we randomly generate customers (with different distributions of origins and destination location), they are assigned to available taxis, and we simulate the movement of taxis to pick up a customer, to drive her to her destination and then waiting for the assignment of a new customer. The simulations are not aimed to be realistic in terms of reflecting the real world operation of a taxi fleet.

Instead, we want to analyze and compare the different assignment strategies proposed in this paper. Thus, we simplified the movements of taxis to straight-line movements with a constant velocity of 17 km/h. This velocity is within the range of the average velocity in the city center of Madrid. That is, we do not take into account neither the real road network, nor the possibility of different traffic conditions.

The general parameters used in the simulation are as follows. We use 1000 taxis (initially distributed randomly in the area with a uniform distribution) and simulate their operation during 5 hours. The taxis do not cruise, that is they only move if they are assigned to a customer. With respect to the customers, we randomly generate a fixed number of customers per each 15 minute interval. In order to analyze different supply/demand ratios, we accomplish different simulation runs with different numbers of customers from 250 to 1000 per 15 minutes, in steps of 125. During the simulation, customers appear at their generated positions and time.

We analyze two different methods to generate the origin (point of appearance) and destination location of customers:

- *Uniform*: both, destinations and origins are generated using a uniform probability distribution over the area of interest
- *Center*: each trip goes either from the outside of the area to the center, or vice versa. The points themselves are generated using a normal distribution (either in the center or in the outside of the area)

The second method corresponds more closely to the actual distributions of taxi trips in urban areas.

We also tested a third distribution, where we defined two density areas that are about 5 km away from each other. Then, each trip is composed of one point (either origin or destination) in one of those areas and the other point is generated with a uniform distribution in the whole region. However, we omit the results for this distribution since they are not significantly different from the *Center* distribution.

In the experiments, we compare the 3 different assignment strategies described in section 3 (without compensations) *FCFS, NTNR* and *FA,* with the three compensation methods *MinDist, MaxRev* and *MinDist/MaxRev*. In the latter cases, algorithm 1 is applied with the described compensation schema. In the *MinDist/MaxRev* approach we apply a scaling factor for monetary incomes of $\gamma= 1/0.00085,$ which corresponds to the net benefit a taxi receives per meter when transporting a customer in the used payment scheme.

During the simulation, the assignment of taxis to (waiting) customers is accomplished every 5 seconds using the corresponding assignment strategy. Once a taxi is assigned to a customer, it moves towards the customers location. After a taxi has reached the location of the assigned customer, it picks up the customer and drives her to her destination. Then, the taxi waits at this point for a new assignment. We simulate fixed pick-up and drop-off times of customers of 30 and 90 seconds, respectively.

The payment scheme we used in the experiments is the one that has been used in the city of Madrid in the last years. A taxi trip has a fixed cost of 2.4 euros and each kilometer a customer moves with the taxi is paid by 1.05 euros. In addition, we assume a cost of operation of a taxi of 0.2 euros per kilometer. This includes petrol, vehicle maintenance cost, as well as other fixed costs.

In the experiments we analyze the average waiting time of customers (the time between the appearance of a customer and of taxis) and revenue of the mediator service.

Each experiment is repeated 10 times with a different random seed, in order to avoid biased results due to a particular distribution of clients. The presented results are averages over those 10 runs.

### 5.1. Unknown customer destinations

In the first set of experiments, we assume that during the assignment phase, the destination of the customers is not known. Thus, from the perspective of a taxi driver, the expected travel time for the ride can be assumed to be some average value for all customers. In our experiments the chosen value has been 4750 meters, roughly the average of all generated taxi trips. This implies that, the only parameter that makes a driver prefer one customer over another is the distance to those customers.

Table 2 and Table 3 show the average customer waiting times for the two different trip distributions (Uniform and Center), respectively. In each case we present the average waiting time of the NTNR method as a baseline result and the absolute and relative variation of waiting times with the other methods.

**Table 2.** Average Waiting times for customers for NTNR and variation over NTNR (Absolut in minutes / relative in %) with the Uniform distribution.

| Method | #Customers per hour | | | | | | |
|---|---|---|---|---|---|---|---|
| | 1000 | 1500 | 2000 | 2500 | 3000 | 3500 | 4000 |
| NTNR | 0.84 | 1.02 | 1.29 | 2.19 | 6.78 | 22.98 | 43.66 |
| FCFS | 0 / 0 | 0 / 0 | 0 / 0 | 0 / 0 | 77.03 / 1136.14 | 102.33 / 445.3 | 121.91 / 279.23 |
| FA | -0.01 / -1.19 | -0.03 / -2.94 | -0.09 / -6.98 | -0.56 / -25.57 | -1.48 / -21.83 | -0.97 / -4.22 | -0.84 / -1.92 |
| MinDist | 0 / 0 | -0.01 / -0.98 | -0.03 / -2.33 | -0.21 / -9.59 | -1.05 / -15.49 | -0.82 / -3.57 | -0.62 / -1.42 |
| MaxRev | 0 / 0 | -0.01 / -0.98 | -0.04 / -3.1 | -0.38 / -17.35 | -1.26 / -18.58 | -0.84 / -3.66 | -0.68 / -1.56 |
| MinDist/MaxRev | 0 / 0 | -0.02 / -1.96 | -0.05 / -3.88 | -0.43 / -19.63 | -1.4 / -20.65 | -0.89 / -3.87 | -0.68 / -1.56 |

**Table 3.** Average Waiting times for customers for NTNR and variation over NTNR (Absolut in minutes / relative in %) with the Center distribution.

| Method | #Customers per hour | | | | | | |
|---|---|---|---|---|---|---|---|
| | 1000 | 1500 | 2000 | 2500 | 3000 | 3500 | 4000 |
| NTNR | 1.2 | 1.56 | 1.92 | 3.83 | 8.29 | 27.38 | 49.01 |
| FCFS | 0 / 0 | 0 / 0 | 0 / 0 | 35.19 / 918.8 | 67 / 808.2 | 84.63 / 309.09 | 99 / 202 |
| FA | -0.08 / -6.67 | -0.21 / -13.46 | -0.36 / -18.75 | -1.95 / -50.91 | -1.88 / -22.68 | -2.42 / -8.84 | -2.67 / -5.45 |
| MinDist | -0.04 / -3.33 | -0.1 / -6.41 | -0.15 / -7.81 | -1.36 / -35.51 | -1.19 / -14.35 | -2.08 / -7.6 | -2.58 / -5.26 |
| MaxRev | -0.04 / -3.33 | -0.16 / -10.26 | -0.29 / -15.1 | -1.81 / -47.26 | -1.75 / -21.11 | -2.45 / -8.95 | -2.74 / -5.59 |
| MinDist/MaxRev | -0.05 / -4.17 | -0.17 / -10.9 | -0.31 / -16.15 | -1.85 / -48.3 | -1.86 / -22.44 | -2.48 / -9.06 | -2.8 / -5.71 |

In both cases, it can be clearly observed that the standard FCFS approach performs considerably worse than the nearest taxi/nearest request method, when the demand increases and the system gets saturated. The saturation point, that is, the number of clients per hour from which the taxis are not able to serve all clients anymore, is between 2000 and 2500 customers per hour in the case of the FCFS method (depending on the spatial distribution of customers). Basically, in a saturation scenario, a good heuristic is to assign taxis in a way that the customers which are closest are served first. In this way, taxis can serve more customers in the same time. This is exactly the basis of the NTNR approach.

The saturation point of all other methods is around 2500-3000 customers per hour. The full auction approach (FA) performs best in almost all cases. This is reasonable since the optimal assignments are calculated among all possible taxis and all possible customers. The obtained improvement with the FA method shows that reducing the overall travel times of all taxis to the nearest customers in each moment produces reductions in the global waiting times of customers. The method, however, is not always the optimal solution when we consider the dynamics of the system over time, e.g., the appearance of future customers. On the long run, it would be better to serve customers first that are within a short distance and also have a destination in an area with a high probability of appearance of new customers. However, in this paper we did not take into account such considerations.

As argued before, the FA method is not applicable in the case of autonomous taxi drivers. In such a case the three different compensation approaches, MinDist, MaxRev, and MinDist/MaxRev can be employed. The three methods outperform the NTNR approach in both customer distributions. The improvements are rather small for low demand scenarios and are very considerably closed to the saturation point (between 2000 and 3000 customers per hour). Above this point, the improvements remain rather stable. Out of the three compensation methods, the approach of maximizing the outcome of the mediator performs better that minimizing the distances to the customers. This is basically due to the fact that some optimal solutions found with the MinDist approach imply high negative revenues for the mediator and, thus, they will not be applied. On the other hand, the MaxRev approach finds the solution with the highest (positive) outcome for the mediator. Since in our settings a positive outcome is only achievable if the sum of the distances of the taxis towards the assigned customers is reduced, a high positive outcome also implies a better solution in terms of reduced distances to customers. This means, in situations where the optimal solution in terms of distances would have a negative outcome for the mediator, MaxRev is able to find good solutions with a positive outcome. In the experiments, the combination of both methods, MinDist/MaxRev, obtains slightly better results in general in comparison to MaxRev.

With regard to the different distributions of customer origin and destination points, the methods show a different behavior with the Uniform distribution in comparison to the Center distribution. Basically, the improvements that can be obtained with smarter assignment solutions than NTNR are lower in the case of the uniform distribution. Our explanation of this fact is that in such a case, also the taxis will always be distributed in an almost uniform way in the region, since they move the customers to their destination locations (which are uniformly distributed). Thus, if both, taxis and customers are distributed in a uniform manner, also the distances between unassigned customers and available taxis will be more homogeneous at all times than in other distributions. This implies that there is not so much space for obtaining improvements when reassigning customers. In general, however, the results in the case of the uniform distribution are similar to the center distribution, even though the obtained improvements are smaller. The FCFS has a clearly worse performance, FA provides the best results and the three compensation approaches obtain improvements with respect to the NTNR approach.

In Figure 3 we analyze the overall revenue of the system, composed of the revenues of the taxis plus the accumulated revenue of the mediator service.

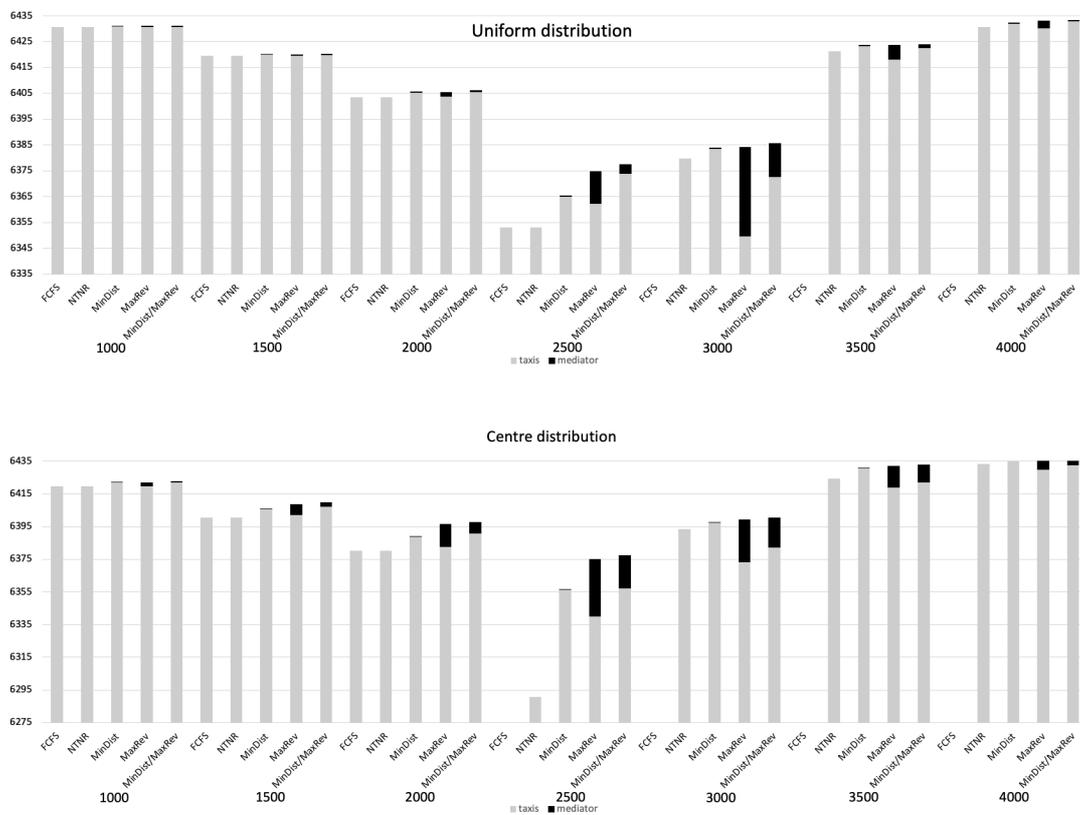

**Figure 3.** Overall revenue of the system in euros for the Uniform distribution and the Center distribution and for different numbers of customers per hour. The outcome is normalized to 1000 customers and is composed of the benefit of the (1000) taxis plus the benefit of the mediator service.

As it can be observed, the highest overall benefit of the system is obtained in all cases with MinDist/MaxRev. It is slightly higher than the other two compensation approaches. And they all outperform the baseline method NTNR. The FCFS approach performs considerably worse than the baseline for higher demand scenarios.

The main difference among the three compensation approaches is the distribution of the benefit among taxis and the mediator service. Taxis have the highest benefit with the MinDist methods. On the other hand, the benefit of the mediator is highest with the MaxRev approach. For instance, in the case of the Center distribution, the mediator benefit for each 1000 customers is between about 2 euros with 1000 customers per hour and around 34 euros with 3000 customers per hour. The combination approach, MinDist/MaxRev provides an intermediate solution, ranging the mediator benefit between 1 and 20 euros in the case of the Center distribution.

An interesting issue is that the average benefit of taxis is higher than in the baseline method for very high and rather low demands. But it might be lower in the case of the MaxRev approach for demands closed to the saturation point. Still, the assignments and reassignments with this method are always done such that a taxi driver does never loose benefit when he is reassigned to another customer. In the case of the MinDist approach, the outcome for taxis is always higher than with the NTNR method. And in the combination method MinDist/MaxRev, taxi benefit is generally higher, except closed to the saturation point, where it is similar or slightly lower. However, it should be noted that, in general, the globally best solution seems to be the MinDist/MaxRev approach. It provides the best customer waiting times, and assures the highest global benefit. Moreover, if the aim of the system is to maximize the revenues of the taxi drivers, the mediator revenues could be simply redistributed among all involved taxis. On the other hand, if the aim of the system is to obtain the highest possible benefit of the mediator service, then the MaxRev approach would be the most appropriate.

## 5.2. Known customer destinations

In the second set of experiments we assume that the customer destinations are known to the taxi drivers when customers are assigned. In this case, the necessary compensation does not only depend on the distance to the customer, but also on the distance required to transport the customer to her destination point. Here, a customer that is closer to the current position of a taxi might not always be a "better" customer, since the paid part of the trip might be much shorter. In Table 4, we present the results of average waiting times with the different methods if the destinations of customers are known when they request a service. We only present the results for the center distribution of customer locations. For the uniform distribution the behavior is very similar.

**Table 4.** Average Waiting times for customers for NTNR and variation over NTNR (Absolut in minutes / relative in %) with the Center distribution if customer destinations are known.

| Method | #Customers per hour | | | | | | |
|---|---|---|---|---|---|---|---|
| | 1000 | 1500 | 2000 | 2500 | 3000 | 3500 | 4000 |
| NTNR | 1.2 | 1.56 | 1.92 | 3.83 | 8.29 | 27.38 | 49.01 |
| FCFS | 0 / 0 | 0 / 0 | 0 / 0 | 35.19 / 918.8 | 67 / 808.2 | 84.63 / 309.09 | 99 / 202 |
| FA | -0.08 / -6.67 | -0.21 / -13.46 | -0.36 / -18.75 | -1.95 / -50.91 | -1.88 / -22.68 | -2.42 / -8.84 | -2.67 / -5.45 |
| MinDist | 0 / 0 | 0 / 0 | 0 / 0 | -0.2 / -5.22 | -0.42 / -5.07 | -1.6 / -5.84 | -2.24 / -4.57 |
| MaxRev | -0.01 / -0.83 | -0.07 / -4.49 | -0.16 / -8.33 | -1.6 / -41.78 | -1.33 / -16.04 | -1.79 / -6.54 | -1.98 / -4.04 |
| MinDist/MaxRev | -0.01 / -0.83 | -0.08 / -5.13 | -0.19 / -9.9 | -1.72 / -44.91 | -1.46 / -17.61 | -2.14 / -7.82 | -2.39 / -4.88 |

As it can be observed in Table 4, the NTNR, FCFS and FA methods do not change if customer destinations are known, since this additional information is not taken into consideration. With regard to the compensation methods, MinDist does not improve the waiting times up to 2500 customers per hour. This is due to the fact that compensations that would have to be paid to taxi drivers in a global reassignment are much higher if customer destinations are known. In our compensation method, taxis never lose money when they are reassigned to a new customer and are compensated if the new customer is "worse". If the customer distance is known, now a taxi would need to be compensated not only for extra distance towards a new customer, but also for the possible loss in the trip with the customer. This implies that most of the time, the mediator would not earn any money with a reassignment. Instead it would have to pay a high compensation to taxis and, thus, would incur in a negative benefit. Since we assume that the mediator should not have a negative overall outcome, in such cases the reassignment would simply not be applied. In the MinDist approach, reassignments with positive outcome for the mediator can only be found for high demand scenarios.

The MaxRev and MinDist/MaxRev methods behave better in this case than MinDist. However, in comparison to the case where customer destinations are not known (and, thus, treated as equal), the results are slightly lower. Again, this is due to the difficulties for finding global reassignment solutions with a positive gain for the mediator.

Figure 4 presents the benefit of the global system in this set of experiments. The MinDist approach behaves very similar to the NTNR method, albeit it assures that taxis always have a slightly higher benefit. The benefit of the mediator is almost zero. MaxRev and MinDist/MaxRev provide the highest overall revenue for the system, being the combination method slightly better in all cases. The mediator benefit is similar as for the case where customer distances are unknown (and estimated through an average value).

MaxRev again assures the highest benefit for the mediator, ranging from about 1 euro and 54 euros for each 1000 customers at a generation rate of 1000 and 3000 customers per hour, respectively.

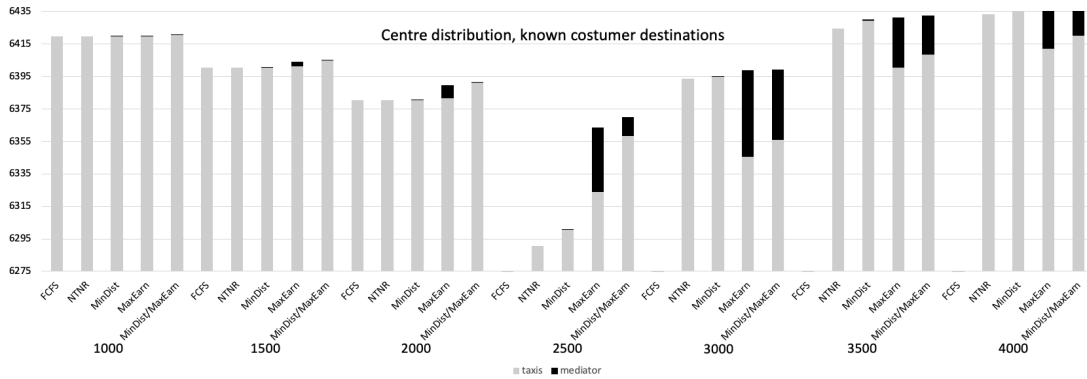

**Figure 4.** Overall revenue of the system in euros for the Center distribution and for different numbers of customers per hour if customer destinations are known. The outcome is normalized to 1000 customers and is composed of the benefit of the (1000) taxis plus the benefit of the mediator service.

# 6. Conclusion

In this paper, we have presented a new algorithm for dynamic taxi dispatching. Our proposal characterizes and differentiates from other approaches in the possibility of proposing taxi reassignments that improve a current global assignment when new customers or available taxis appear. In order to create a realistic dispatching proposal that is accepted by rational autonomous taxi drivers, we devised a compensation system in which taxis that get a "worse" customer receive an extra monetary compensation while those who get "better" customers pay part of the reduced costs to a mediator service. The mediator service (e.g. a taxi company), manages all those payments and only proposes a better assignment if its accumulated revenue is positive and thus, it does not incur in extra costs. We evaluated three different versions of our proposal, namely (i) minimizing distances, (ii) maximizing mediator revenue, and (iii) a combination of both. We compared them to the standard FCFS dispatching strategy and its modified NTNR approach, as well as to a "complete" dynamic reassignment approach without compensations (FA). The results showed that our methods outperform the standard strategies, especially when the demand increases. It obtains very similar results as the FA approach.

A key lesson learnt from our experiments is that, especially in high load situations, a dynamic reassignment approach can produce noticeable benefits for system performance. This is true for all stakeholders: our proposal contributes to reduce average waiting time of customers and also helps to increase drivers' revenues. In some configurations, the new assignments even allow obtaining some economic benefit for the mediator as well, resulting from exceeding incomes of compensations. Such extra money could be shared, for instance, to the taxis. Notice that the performance of the dispatching strategy could be improved even further if the compensation system allowed a negative balance. This would not be necessarily unrealistic if the mediator was a public entity. For example, a municipality might be willing to invest some money if a more efficient service is provided, thus reducing $CO_2$ emissions (probably reducing city fines for high emissions imposed by superior authorities).

Still, while we have obtained promising results, it should be noticed that our approach does rely on some simplifications. Obviously, assuming that travel time is proportional to distance is one of them. Furthermore, we assumed that all taxi drivers are (economically) rational utility maximisers, and that even a very small additional monetary benefit can make them accept changing their plans, which may not always be true. Finally, we do not consider the possibility of taxi drivers to "opt out" of our mechanism, e.g. by not accepting its indications (for whatever reason). This may introduce additional "noise" in our assignments with a potentially negative impact on their performance.

The taxi dispatching model presented in this article opens up several avenues for subsequent research in the field of knowledge-based and expert systems. Our future work will unfold among several lines. Firstly, we intend to relax some of the simplifying assumptions that our current approach relies on. In particular, we will base the different cost functions used in our mechanism on a more accurate estimation of travel times and distances. For this purpose, we will set out from a realistic road network topology and real-world load data. This information is already available to us for the town of Madrid. We will then be able to perform our simulations with a fully-fledged microscopic traffic simulator such as SUMO[5]. In this context, we plan to reach an agreement with a taxi fleet operator so as to use real-world data on customer origin and destinations for the experiments.

Secondly, similar to the problems addressed by the works of Laha and Putatunda (2018) or Ocalir et al. (2010), we plan to apply methods to estimate probabilities of new taxi customers appearing in different areas of the network. In a previous paper related to the dynamic positioning of ambulances, we used Voronoi Tessellation to dynamically determine the default positions for service vehicles (Billhardt et al., 2014) based on historical data. Such information can be complemented with domain knowledge of a particular destination (soccer matches, concerts, etc.). The corresponding information would then be used to make the taxi dispatch decisions more predictive.

Thirdly, in line with the work by Massowa y Canbolat (2010), we intend to look into more complex models of the taxi drivers' reactions to our dispatch strategy, so as to better adjust compensations and provide the designer of a taxi dispatching model with guidelines to choose among alternative strategies for revenue distribution (MinDist, MaxRev, etc.). To this respect, we plan to analyze how the ratio of taxi drivers that decide to refrain from using our dispatching service affects to the global performance of the system.

Fourthly, in this context, we also plan to look into coalition formation techniques from Cooperative Game Theory. In general, algorithms to compute solutions to coalitional games are known to suffer complexity problems, but we have applied an approximation scheme (the bilateral Shapely Value) in the context of a Smart Grid application (Mihailescu et al., 2017), and would like to explore as to how far a fair and efficient split of the savings implied by taxi reassignment can be implemented in this manner.

Finally, it should be noted that the mechanism proposed in this article does not only apply to taxi fleets but could also be adapted to other types of *open fleets* (Billhardt et al., 2017), where autonomous drivers with individual objectives provide some transportation service (e.g., messenger or parcel delivery services). In particular, to this respect, we will investigate the use of heterogeneous fleets or "CyberFleets", as proposed in (Billhardt et al. 2014b). The proposal from this article could even provide a suitable basis for managing other sorts of large-scale Demand Responsive Transport systems that provide shared transportation services with flexible routes (Satunin and Babkin, 2014).

## Acknowledgements

This work was supported by the Autonomous Region of Madrid (grant "MOSI-AGIL-CM" (S2013/ICE-3019) co-funded by EU Structural Funds FSE and FEDER), project "SURF" (TIN2015-65515-C4-X-R (MINECO /FEDER)) funded by the Spanish Ministry of Economy and Competitiveness, and through the Excellence Research Group GES2ME (Ref. 30VCPIGI05) co-funded by URJC and Santander Bank.

---

[5] http://sumo.dlr.de/index.html